%% file: denoising.tex
\documentclass[conference]{IEEEtran}
\IEEEoverridecommandlockouts
\usepackage{ifpdf}
\usepackage{pgfplots}
\usepackage{tikz}
\usepackage{subfig}                     % for subfigures
\usepackage{float}
\usepackage{booktabs}                   % nice looking tables (for tables with ONLY horizontal lines)

\usepackage[T1]{fontenc}
\usepackage[utf8]{inputenc}
\usepackage[utf8]{inputenc}
\usepackage{pgfplots}
\usepackage{grffile}
\pgfplotsset{compat=newest}
\usetikzlibrary{plotmarks}
\usetikzlibrary{arrows.meta}
\usepgfplotslibrary{patchplots}

\usepackage{cite}
\usepackage{amsmath,amssymb,amsfonts}
\usepackage{algorithmic}
\usepackage{graphicx}
\usepackage{adjustbox}
\usepackage{units}
\usepackage{textcomp}
\usepackage{xcolor}
\usepackage{stfloats}
\usepackage{epstopdf}
\usepackage{bm}
\usepackage{amssymb}
\usepackage[abs]{overpic}
\definecolor{mycolor1}{rgb}{0.14902,0.14902,0.14902}%
\definecolor{mycolor2}{rgb}{0.72941,0.83137,0.95686}%
\def\BibTeX{{\rm B\kern-.05em{\sc i\kern-.025em b}\kern-.08em
    T\kern-.1667em\lower.7ex\hbox{E}\kern-.125emX}}
\renewcommand{\baselinestretch}{0.97}
\begin{document}

\title{Towards Adversarial Denoising of Radar Micro-Doppler Signatures}
%	\title{An Analytical Study of Human Characteristics on Micro-Doppler Identification using Deep Learning}
%	\title{An Analytical Study of Human Identification based on Micro-Doppler Signatures using Deep Learning}

%\author{\IEEEauthorblockN{Sherif Abdulatif}
%\IEEEauthorblockA{Institute of Biomechatronic Systems\\
%Fraunhofer Institute for Manufacturing Engineering and Automation IPA\\
%Noberlstr. 12, Stuttgart 70569\\
%Email: sherif.abdulatif@ipa.fraunhofer.de}
%\and
%\IEEEauthorblockN{Homer Simpson}
%\IEEEauthorblockA{Twentieth Century Fox\\
%\and
%\IEEEauthorblockN{James Kirk\\ and Montgomery Scott}
%\IEEEauthorblockA{Starfleet Academy\\
%San Francisco, California 96678-2391\\


\author{\IEEEauthorblockN{Sherif Abdulatif\IEEEauthorrefmark{4}\IEEEauthorrefmark{1},
		Karim Armanious\IEEEauthorrefmark{4}\IEEEauthorrefmark{1},
		Fady Aziz\IEEEauthorrefmark{2},
		Urs Schneider\IEEEauthorrefmark{2},
		Bin Yang\IEEEauthorrefmark{4}}
	\IEEEauthorblockA{\IEEEauthorrefmark{4}Institute of Signal Processing and System Theory,
		University of Stuttgart\\
		\IEEEauthorrefmark{2}Fraunhofer Institute for Manufacturing Engineering and Automation IPA \\
		Email: \{sherif.abdulatif, karim.armanious\}@iss.uni-stuttgart.de\\
		\IEEEauthorrefmark{1}These authors contributed to this work equally.}}


\maketitle

\begin{abstract}
	%\boldmath
	Generative Adversarial Networks (GANs) are considered the state-of-the-art in the field of image generation. They learn the joint distribution of the training data and attempt to generate new data samples in high dimensional space following the same distribution as the input. Recent improvements in GANs opened the field to many other computer vision applications based on improving and changing the characteristics of the input image to follow some given training requirements. In this paper, we propose a novel technique for the denoising and reconstruction of the micro-Doppler ($\boldsymbol{\mu}$-D) spectra of walking humans based on GANs. Two sets of experiments were collected on 22 subjects walking on a treadmill at an intermediate velocity using a \unit[25]{GHz} CW radar. In one set, a clean $\boldsymbol{\mu}$-D spectrum is collected for each subject by placing the radar at a close distance to the subject. In the other set, variations are introduced in the experiment setup to introduce different noise and clutter effects on the spectrum by changing the distance and placing reflective objects between the radar and the target. Synthetic paired noisy and noise-free spectra were used for training, while validation was carried out on the real noisy measured data. Finally, qualitative and quantitative comparison with other classical radar denoising approaches in the literature demonstrated the proposed GANs framework is better and more robust to different noise levels. 
	%		Radar sensors can be used for analyzing the induced frequency shifts due to micro-motions in both range and velocity dimensions identified as micro-Doppler ($\boldsymbol{\mu}$-D) and micro-Range ($\boldsymbol{\mu}$-R), respectively. Different moving targets will have unique $\boldsymbol{\mu}$-D and $\boldsymbol{\mu}$-R signatures that can be used for target classification. Such classification can be used in numerous fields, such as gait recognition, safety and surveillance. In this paper, a \unit[25]{GHz} FMCW Single-Input Single-Output (SISO) radar is used in industrial safety for real-time human-robot identification. Due to the real-time constraint, joint Range-Doppler (R-D) maps are directly analyzed for our classification problem. Furthermore, a comparison between the conventional classical learning approaches with handcrafted extracted features, ensemble classifiers and deep learning approaches is presented. For ensemble classifiers, restructured range and velocity profiles are passed directly to ensemble trees, such as gradient boosting and random forest without feature extraction. Finally, a Deep Convolutional Neural Network (DCNN) is used and raw R-D images are directly fed into the constructed network. DCNN shows a superior performance of 99\% accuracy in identifying humans from robots on a single R-D map.  
	
\end{abstract}


% \fi
\IEEEpeerreviewmaketitle

\vspace{-0.3cm}
\section{Introduction \label{sec:int}}
Radar sensors are widely used nowadays in the tracking and identification of different targets. This is due to its superior penetration capabilities and robustness against different lighting and weather conditions in comparison to vision-based approaches such as camera and LIDAR \cite{rasshofer2011influences,wang2011integrating}. Micro-Doppler ($\boldsymbol{\mu}$-D) spectrum introduces an additional capability in radar systems by allowing the analysis of the individual micro motions of moving targets. For instance, when studying the $\boldsymbol{\mu}$-D signature of a non-rigid body as a flying bird, some frequencies reflect the translational velocity of the bird and other frequencies reflect the wings motion. The same holds for a helicopter where the body indicates the relative velocity and the periodic frequencies reflect the blades movement \cite{microdopplerBook}. Accordingly, $\boldsymbol{\mu}$-D signatures proved to be an effective measure of human locomotions such as swinging of body limbs. 

Although some human motions are periodic, e.g. walking and running, in general each human body part has a micro motion which induces a different $\boldsymbol{\mu}$-D signature. Based on the study presented in \cite{abdulatif2017radarconf}, the $\boldsymbol{\mu}$-D signature of human activity is analyzed as the superposition of different limbs signatures involved in such activity. Accordingly, the superpositioned $\boldsymbol{\mu}$-D signature is used for differentiating between different periodic and aperiodic human activities. Previous studies inspected the use of $\boldsymbol{\mu}$-Ds in fall detection especially for monitoring elderly people \cite{moeness2016fall,moeness2018fall}. Moreover, the  $\boldsymbol{\mu}$-D capability in radar is widely used for rehabilitation tasks to monitor gait abnormalities and caned assisted walking \cite{ann2017radarconf}. Additionally, radar is nowadays the dominant sensor used for autonomous driving as it combines high resolution in range, velocity and depth perception. This motivated the use of radar in vital tasks such as on-road pedestrians and cyclists recognition \cite{wagner2017pedestrian}.

However, most of the previous research assumes perfect measurement conditions where an ideal obstacle-free environment is preserved. This is handled by placing the radar at a suitable distance above the ground to maintain a clear line of sight with the target of interest \cite{moeness2016fall}. Such requirements are not always applicable, and in many realistic cases, an obstacle-free environment is not guaranteed. Moreover, the distance between the radar and the target of interest is directly related to the measurement Signal-Noise-Ratio (SNR). The larger this distance is, the smaller the detected Radar Cross Section (RCS), which leads to a lower SNR \cite{richards2005fundamentals}. It is explained in \cite{Chen2010,li2013geoscience,fugui2017clutter} that a significant degradation in the classification and detection performance is observed in low SNR cases. To overcome such degradation, $\boldsymbol{\mu}$-D spectrum denoising techniques were introduced to allow for scene-independent radar performance in non-ideal measurement conditions. 

In \cite{du2013clean}, authors proposed the use of CLEAN decomposition technique for noise reduction and subsequent enhanced classification of aeroplane targets. The main limitation of this approach is the requirement of accurate prior knowledge of the measurement SNR level, which is not always practically feasible. To overcome this limitation, authors in \cite{du2015CPPCA} used Complex Probabilistic Principal  Component  Analysis (CPPCA) to reconstruct the radar spectrum and then discard the noise in latent space. Furthermore, classical adaptive thresholding approaches such as Constant False Alarm Rate (CFAR) \cite{kronauge2013cfar2d}, which is extensively used for range estimation and clutter suppression, can be adopted for $\boldsymbol{\mu}$-D denoising as mentioned in \cite{foued2017cfar}. Although this technique results in an enhanced classification performance of the denoised spectrum, it requires long inference time per spectrum due to high computational complexity. Finally, image-based denoising approaches such as Gamma-correction can also be used for $\boldsymbol{\mu}$-D spectrum denoising \cite{poynton1998rehabilitation}. Gamma-correction can be considered as a simple but effective non-linear mapping technique where a constant value $\gamma$ is used to map each pixel intensity $I$ to a corrected pixel intensity $\hat{I}=I^\gamma$. According to the chosen value of $\gamma$, the background noise is suppressed while maintaining the overall spectrum structure. Conversely, Gamma-correction may result in the adverse elimination of low power vital components within the spectrum. This may affect further applications which require knowledge of the local components as well as the overall spectrum structure, such as activity recognition and human subject identification \cite{park2016micro,kim2016activity,cao2018id}. Moreover, all of the mentioned denoising techniques are specific to a fixed measurement scenario or SNR level. Generalization to realistically diverse structures requires case-specific manual tuning which hinders the denoising automation capability.

In this work, we propose the use of state-of-the-art deep learning techniques for adaptive denoising and reconstruction of $\boldsymbol{\mu}$-D spectrum. To this end, we propose the use of Generative Adversarial Networks (GANs) introduced in 2014 \cite{ian2014GAN}. GANs consists of two Deep Convolutional Neural Networks (DCNNs) trained in competition with each other. The first network is a generator aiming to generate samples from the same distribution as the input samples. The second network is a discriminator aiming to discriminate the fake generator output samples from the real input samples. Both networks are trained with competing objectives, the generator trying to fool the discriminator and the discriminator is trained to improve its classification performance to avoid being fooled. For image-to-image translation tasks, other variants of GANs based on conditional GAN (cGAN) are used \cite{mirza2O14cgan}. In recent years, a wide variety of image-to-image applications are making use of cGANs in different fields, such as motion correction, image inpainting and super-resolution tasks \cite{armanious2018medgan,miyato2018super,armanious2018inpainting}.

The main goal of this work is to introduce the use of cGAN as a $\boldsymbol{\mu}$-D denoising technique which is generalizable to different SNR levels and measurement scenarios with fast inference time. Additionally, the proposed method must maintain both the global and the local spectrum structure to preserve the performance of post-processing tasks. We illustrate the advantages of the proposed denoising method by qualitative and quantitative comparisons with classical denoising techniques.



%\hfill November 1, 2016
\vspace{-2mm}
\section{Methodology}
\begin{figure*}
	\centering
	\includegraphics[width=1\textwidth]{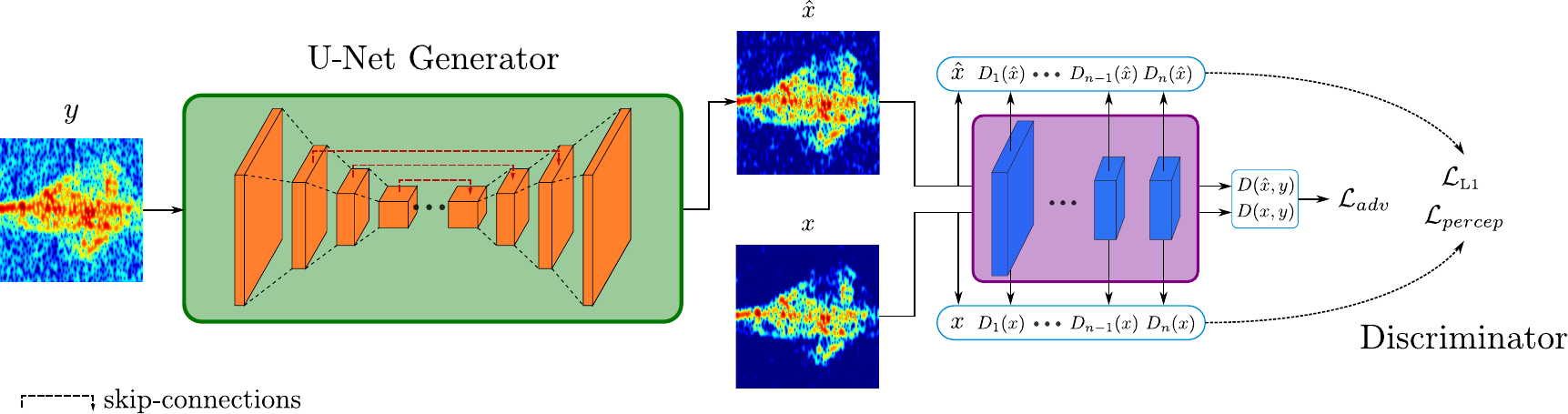}
	\caption{An overview of the proposed architecture with non-adversarial losses for $\boldsymbol{\mu}$-D spectrum denoising and reconstruction.\label{fig:netArch}}
	\vspace{-5mm}
\end{figure*}
The proposed denoising and reconstruction network architecture is based on cGANs. However, additional non-adversarial losses are used with the cGAN structure to preserve the details of the denoised spectrum. An overview of the proposed architecture is depicted in Fig.~\ref{fig:netArch}. 
\vspace{-2mm}
\subsection{Conditional Generative Adversarial Networks \label{sec:cgan}}
In contrast to traditional GANs, a conditional generator is trained to map a source domain sample, in our case the noisy radar $\boldsymbol{\mu}$-D spectrum $y$, to a generated sample $\hat{x}=G(y)$ representing the corresponding denoised spectrum. The generated sample $\hat{x}$ is then passed as an input to a discriminator network along with the corresponding paired ground truth $x$ which is aligned with the source domain input $y$. The discriminator is trained as a binary classifier to differentiate between the fake sample $\hat{x}$ and the real sample $x$, such that $D(\hat{x},y)=0$ and $D(x,y)=1$. The problem can be illustrated as a minimization-maximization problem by which a generator is generating samples to maximize the probability of fooling the discriminator. At the same time, the discriminator is classifying the real input samples from the generated ones by minimizing the binary cross entropy loss. This part represents the network adversarial loss and the training procedure is expressed as the following min-max optimization problem: 
$$\min_G \max_D \mathcal{L}_{adv}=\mathbb{E}_{x,y}\left[ \log D(x,y)\right] + \mathbb{E}_{\hat{x},y}\left[ (1-\log D(\hat{x},y))\right] $$
\vspace{-8mm}
\subsection{Non-adversarial Losses}
Depending only on the adversarial loss may produce non-realistic spectra with different global structures than the required clean target spectra. To resolve this issue, non-adversarial losses are also introduced as additional losses to insure consistent and realistic spectrum reconstruction. The first non-adversarial loss utilized in the proposed network is the L1 pixel reconstruction loss, which was first introduced as adversarial setting in \cite{isola2016L1}. The L1 loss $\mathcal{L}_{L1}$ is described as the pixel-wise Mean Absolute Error (MAE) between the reconstructed spectrum $\hat{x}$ and the target clean spectrum $x$:
$$\mathcal{L}_{L1}=\mathbb{E}_{x,\hat{x}}\left[ \left\| x-\hat{x} \right\|_1 \right]$$

In addition to the L1 loss, the perceptual loss is utilized to minimize the perceptual discrepancy between the generated output $\hat{x}$ and the corresponding ground truth sample $x$, thus modelling a better representation from the human judgement perspective \cite{johnson2016perceptual}. The perceptual loss considers the output of the intermediate convolutional layers of the discriminator, which represents the extracted features of the input image. Accordingly, the perceptual loss of the $i^{th}$ layer is calculated as the MAE between the extracted features $D_i$ of the desired target image $x$ versus the generator output image $\hat{x}$:
$$F_i(x,\hat{x})=\left\| D_i(x,y)-D_i(\hat{x},y) \right\|_1$$ 
\vspace{-1mm}
During training, the task is to minimize the weighted sum of the perceptual losses of different discriminator layers:
$$\mathcal{L}_{perc}=\sum_{i=1}^{L}\lambda_i F_i(x,\hat{x})$$ where $\lambda_i>0$ is the weight corresponding to the $i^{th}$ layer and $L$ is the index of the last layer in the discriminator network. Hence, the L1 loss can be considered as the perceptual loss of the raw images at the input layer ($i=0$).
\vspace{-2mm}
\subsection{Network Architecture}
As shown in Fig.~\ref{fig:netArch}, the proposed network utilizes the U-Net presented in \cite{isola2016L1} as the generator part, which is typically a convolutional encoder-decoder structure. The task of the U-Net generator is to map a high dimensional source domain input image $y$ (256$\times$256$\times$3) to an output image $\hat{x}$ of the same dimensions representing the same underlying structure but with different image characteristics (translation from noisy to noise-free). In the encoder branch, the image passes through a sequence of convolutional layers with varying number of filters and a 4$\times$4 kernel. In all layers a stride of 2 is used, thus the dimensions of the image is progressively halved until a final latent space representation is reached. The decoder branch is a mirror of the encoder branch, but with upsampling deconvolutional layers to reach the original input dimension. To avoid information loss due to the bottleneck layer, skip connections are used between encoder and corresponding decoder layers to pass vital information between the two branches and ensure a consistent high quality output.

As for the discriminator part, a patch discriminator architecture is utilized. In our proposed architecture the image is divided into $70\times70$ overlapping patches and each patch is classified by the discriminator as real or fake and an average score of all patches is used. Finally, the overall network loss is the sum of all mentioned losses and can be expressed as follows:
$$\mathcal{L}=\mathcal{L}_{adv}+\lambda_{L1}\mathcal{L}_{L1}+\lambda_{perc}\mathcal{L}_{perc}$$ where $\lambda_{L1}$ and $\lambda_{perc}$ are the weights for the L1 and the perceptual losses, respectively. These weights are used to balance the share of all losses by rescaling each loss with an appropriate corresponding value. After extensive hyperparameter optimization, the used weights were set $\lambda_{perc}=0.0001$ and $\lambda_{L1}=100$.
\section{Experimental Setup}
In this study, the proposed denoising method is applied only on the $\boldsymbol{\mu}$-D spectrum of walking human subjects. However, the idea can be generalized to different activities. Walking is particularly selected for easier dataset collection. Furthermore, walking $\boldsymbol{\mu}$-D signatures are challenging as they contain vital information about the walking style, and thus subject identity, in lower-powered regions representing individual limb swings. The model must learn to preserve such information eliminating the surrounding high-powered noise signals.
\vspace{-1mm}
\subsection{Radar Parametrization}
Since only the $\boldsymbol{\mu}$-D spectrum is of interest in this paper, a CW radar is used for data collection. The used radar operates at a carrier frequency $f_c =$ \unit[25]{GHz} with a pulse repetition frequency of $F_p =$ \unit[4]{kHz}. Based on equations presented in \cite{microdopplerBook}, the measurable maximum velocity is $v_{max}\approx$ \unit[12]{m/s}. By definition, a $\boldsymbol{\mu}$-D spectrum is interpreted as a Time Frequency (TF) analysis of the received sampled complex time domain data $x(n)$. In our case, the velocity of the target of interest is not constant, thus the corresponding reflected time domain signal is non-stationary. To visualize the temporal velocity behavior, a spectrogram is calculated as the logarithmic squared magnitude of the Short-Time Fourier Transform (STFT):
$$S(m,f) = 20\log\left\|\sum_{n=-\infty}^{\infty}w(n-m)\,x(n)\,\exp(-j2\pi fn)\right\|$$ where $n$ is the discrete time domain index, $w(.)$ is a smoothing window, and $m$ is the sliding index. A Gaussian smoothing function is used with a window size of 512 to achieve a velocity resolution of $v_{res}\approx$ \unit[0.04]{m/s}.
\vspace{-1mm}
\subsection{Dataset Preparation}
\begin{figure}
	\vspace{-5mm}
	\centering
	\centerline{\hspace*{2mm}\subfloat[Clean experiment setup.]{\includegraphics[scale=.38]{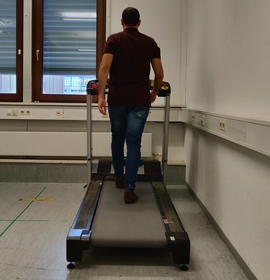}\label{fig:clExp}}
		\hspace{1mm}
		\subfloat[Clean $\boldsymbol{\mu}$-D spectrum.]{\resizebox{.60\columnwidth}{!}{\input{fig/cleanMD_new.tex}}\label{fig:clMD}}}
	\vspace{-2mm}
	\centering
	\centerline{\hspace*{2mm}\subfloat[Noisy experiment setup.]{\includegraphics[scale=.38]{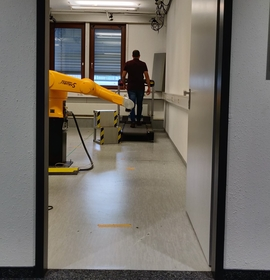}\label{fig:nExp}}
		\hspace{1mm}
		\subfloat[Noisy $\boldsymbol{\mu}$-D specturm.]{\resizebox{.60\columnwidth}{!}{\input{fig/noisyMD_new.tex}}\label{fig:nMD}}}
	\caption{Examples of the setup for both clean and noisy experiments with their corresponding $\boldsymbol{\mu}$-Ds of one full gait cycle. Since the human position is not changing from the radar perspective, a bulk torso velocity of \unit[0]{m/s} is observed. \label{fig:exps}}
	\vspace{-6mm}
\end{figure}
To collect a reasonable amount of data with realistic variations, 22 subjects of different genders, weights, and heights were asked to walk on a treadmill at a velocity of \unit[1.4\,-1.6]{m/s}. The main motivation behind the use of a treadmill is to enforce a motion style for all subjects. Moreover, this mitigates the unwanted variations in radar-target alignment, which scales the measured radial velocity \cite{Bartsch2012Pedestrian} and this is not desired in this paper. As explained in \cite{boulic1990global,umberger2010stance}, a walking human gait cycle is divided into two main phases. A swinging phase where the upper body limbs and lower body limbs are swinging in an alternative behavior (left arm with right foot and vice versa). In the stance phase, both feet are touching the ground and no swinging behavior is observed, only the bulk of the relative velocity. Hence, a full gait cycle can be analyzed as two swinging phases corresponding to the left and right feet swings interrupted with a middle stance phase. 

The proposed cGAN model, presented in Fig.~\ref{fig:netArch}, takes a source domain noisy spectrum and a target clean spectrum during training to learn the spectrum patterns, which will enable spectrum denoising. Accordingly, the first experiment collects the clean target spectrum by placing the radar at a \unit[3]{m} distance facing the back of the treadmill and a target is moving away from the radar with a direct line of sight. The second experiment models noisy spectra, where the radar is placed at a \unit[10]{m} distance from the treadmill. As mentioned in Sec.~\ref{sec:int}, a longer distance results in a lower SNR. Moreover, to include different clutter and fading effects, reflective obstacles were placed between the target and radar. To collect radar reflections with different SNRs, the noisy experiment setup is changed from one subject to another by varying the number of reflective objects between the target and the radar and adapting the radar-target distance from 10m to 8m. Finally, each subject is asked to walk on the treadmill for 3 minutes for each experiment. An example of the setup for both experiments with the corresponding full gait $\boldsymbol{\mu}$-D is depicted in Fig.~\ref{fig:exps}.

\begin{figure*}
	\begin{minipage}[t]{1.0\linewidth}
		\centering
		\begin{overpic}[width=0.18\textwidth]%
			{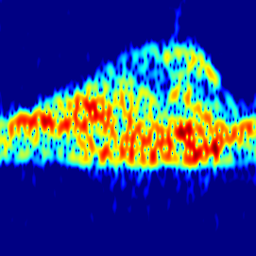}
			\centering
			\put(16,98){Clean spectrum}
		\end{overpic}
		\begin{overpic}[width=0.18\textwidth]%
			{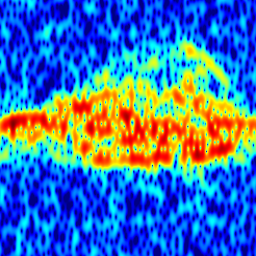}
			\centering
			\put(6,98){Real noisy spectrum}
		\end{overpic}
		\begin{overpic}[width=0.18\textwidth]%
			{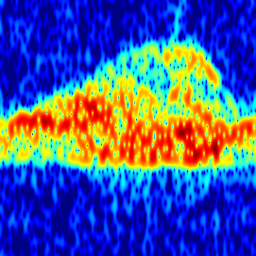}
			\centering
			\put(22,98){SNR 10 dB}
		\end{overpic}
		\begin{overpic}[width=0.18\textwidth]%
			{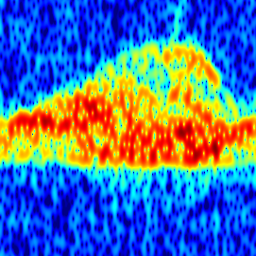}
			\centering
			\put(25,98){SNR 5 dB}
		\end{overpic}
		\begin{overpic}[width=0.18\textwidth]%
			{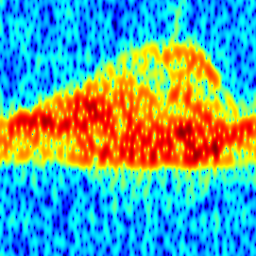}
			\centering
			\put(25,98){SNR 0 dB}
		\end{overpic}		
	\end{minipage}%
	\caption{An example of a clean $\boldsymbol{\mu}$-D spectrum with AWGN at different SNR levels in comparison to the real noisy spectrum.\label{fig:awgn}}
	\vspace{-5mm}
\end{figure*}

To avoid power fading effect during the experiment, a normalization is applied on each gait cycle. It can be observed from Fig.~\ref{fig:clMD} that the two swinging phases are not perfectly symmetric in one full gait cycle, thus the denoising will be applied on half gait basis. This will add more generalization capability to the network regardless of the initial motion gait direction, e.g. left-right in comparison to right-left gaits. Finally, each half gait spectrum will be sliced and resized to an RGB unsigned integer-16 image of a resolution (256$\times$256$\times$3). The duration of a half gait cycle is different from one subject to another and it is highly affected by the step length, which is directly proportional to the height of the subject under test. Based on the proposed treadmill velocities, a half gait cycle duration is analyzed for the 22 subjects and found to be within \unit[0.4\,-\,0.6]{s} interval. On average the total number of images per experiment is 22$\times$180/0.5$\,\approx\,$8000 images.

Since both experiments are done with one radar at different time instants and under different conditions, the dataset of both experiments is unpaired. As mentioned in Sec.~\ref{sec:cgan}, cGANs basically work on paired images in the source and target domain. To resolve this issue, the noise distribution in the noisy $\boldsymbol{\mu}$-D spectra must be modelled and added to the clean target spectra. Then both the clean data and the modelled noise data are passed to the proposed cGAN network in pairs for training on specific subjects. However, the network performance will be validated on real noisy spectra and not the modelled ones. For the first feasibility study, an Additive White Gaussian Noise (AWGN) is added to the logarithm of the  clean target spectra. To make sure that the modeled noise is covering all scenarios in the real experiments, different SNR levels are used. This will improve the generalization performance of the network. As shown in Fig.~\ref{fig:awgn}, the clean and real noisy $\boldsymbol{\mu}$-D are not perfectly aligned. However, using AWGN with different SNR levels will resolve this alignment issue. To validate the AWGN model assumption, a noise analysis is applied to the acquired noisy dataset. In the analysis, each collected noisy half gait $\boldsymbol{\mu}$-D spectrum (dB-scale) is subtracted from a clean $\boldsymbol{\mu}$-D spectrum from the same subject and same swing orientation (right or left swing). This is applied on all 22 collected. Then the error distribution is analyzed. In Fig.~\ref{fig:nAnalysis}, an approximate bell-shape distribution is observed, which validates our AWGN assumption. In Fig.~\ref{fig:nAnalysis}, a certain non-zero mean can be observed. This is due to the misalignment effect discussed above between both clean and noisy dataset.
\begin{figure}
	\centering
	\resizebox{.9\columnwidth}{!}{
		\input{fig/noiseAnalysis.tex}
	}
	\caption{The probability density function of noisy $\boldsymbol{\mu}$-D spectra. \label{fig:nAnalysis}}
	\vspace{-10mm}
\end{figure}
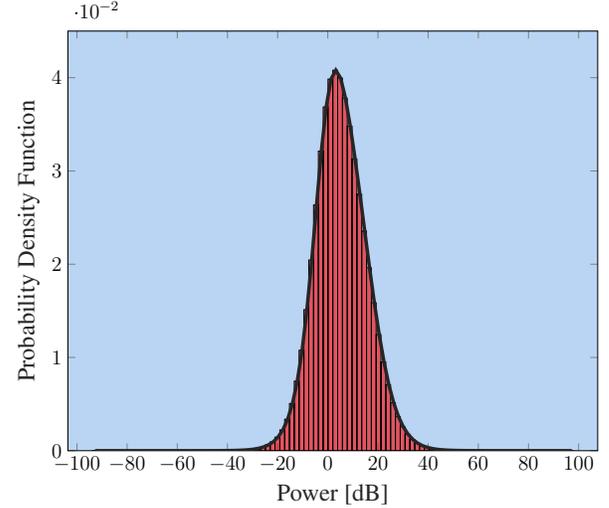
\begin{figure*}[t]
	\begin{minipage}[t]{1.0\linewidth}
		\centering
		\vspace{7mm}
		\begin{minipage}[t]{0.190\linewidth}
			\centering
			\begin{overpic}[width=0.945\textwidth]%
				{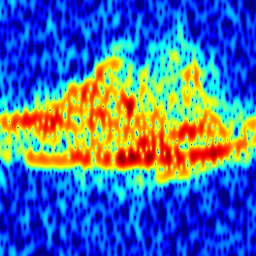}
				\centering
				\put(15,98){Noise-Corrupted}
			\end{overpic}	
		\end{minipage}%
		\begin{minipage}[t]{0.6\linewidth}
			\centering
			\begin{overpic}[width=0.298\textwidth]%
				{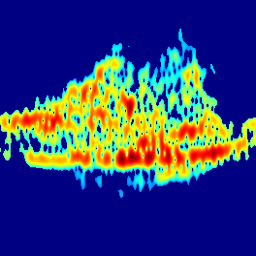}
				\centering
				\put(34,98){CFAR}
			\end{overpic}
			\begin{overpic}[width=0.298\textwidth]%
				{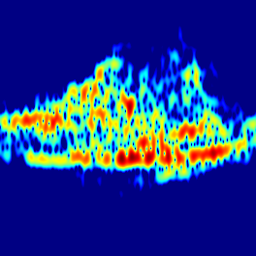}
				\centering
				\put(10,98){Gamma-correction}
			\end{overpic}
			\begin{overpic}[width=0.298\textwidth]%
				{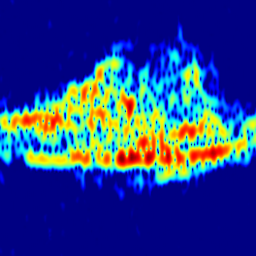}
				\centering
				\put(32,98){cGAN}
			\end{overpic}
		\end{minipage}
		\begin{minipage}[t]{0.190\linewidth}
			\centering
			\begin{overpic}[width=0.945\textwidth]%
				{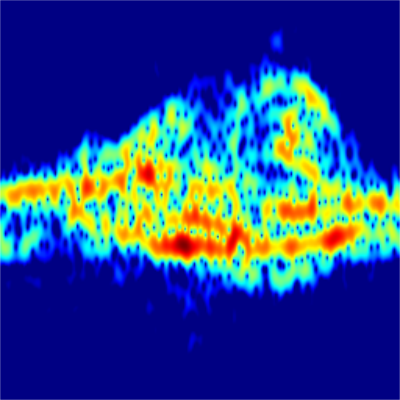}
				\centering
				\put(18,98){Clean spectrum}
			\end{overpic}		
		\end{minipage}\\
		\vspace{2mm} %%%%%%%%%%%%%%%%%%%%%%%%%%%%%%%%%%%%%%%%%%%%%%%%%%%%%%%%%%%%%%%%%%%%%%%
		\begin{minipage}[t]{0.19\linewidth}
			\centering
			\begin{overpic}[width=0.945\textwidth]%
				{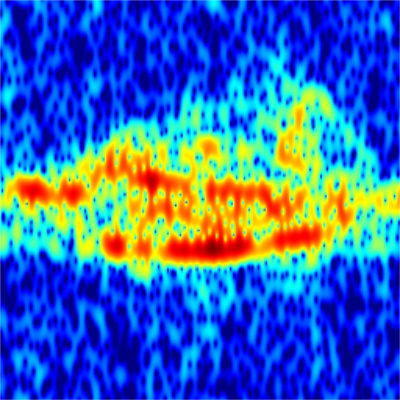}
			\end{overpic}	
		\end{minipage}%
		\begin{minipage}[t]{0.6\linewidth}
			\centering
			\begin{overpic}[width=0.298\textwidth]%
				{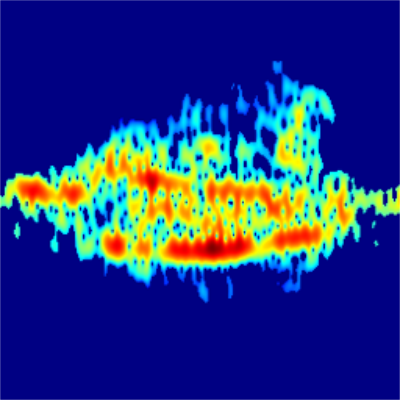}
			\end{overpic}
			\begin{overpic}[width=0.298\textwidth]%
				{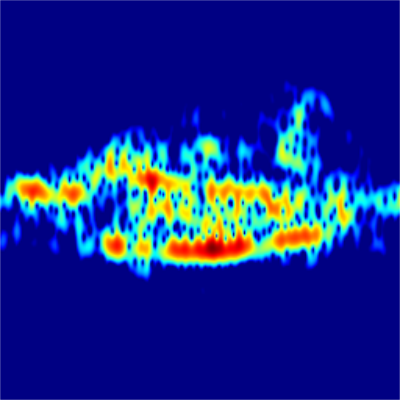}
			\end{overpic}
			\begin{overpic}[width=0.298\textwidth]%
				{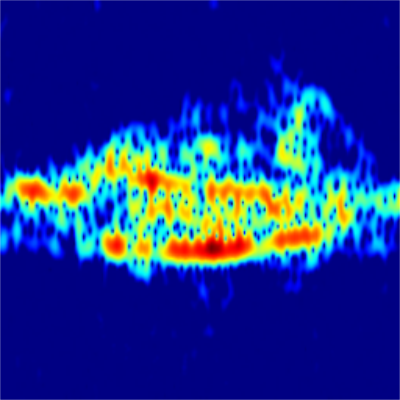}
			\end{overpic}
		\end{minipage}
		\begin{minipage}[t]{0.190\linewidth}
			\centering
			\begin{overpic}[width=0.945\textwidth]%
				{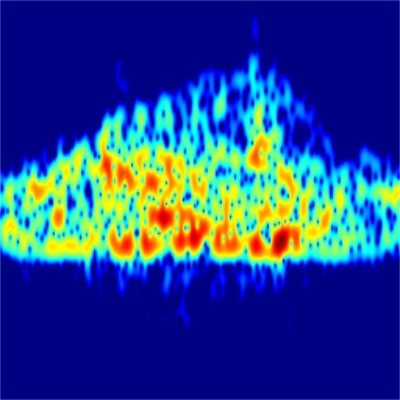}
			\end{overpic}		
		\end{minipage}\\
		\vspace{2mm} %%%%%%%%%%%%%%%%%%%%%%%%%%%%%%%%%%%%%%%%%%%%%%%%%%%%%%%%%%%%%%%%%%%%%%%
		\begin{minipage}[t]{0.19\linewidth}
			\centering
			\begin{overpic}[width=0.945\textwidth]%
				{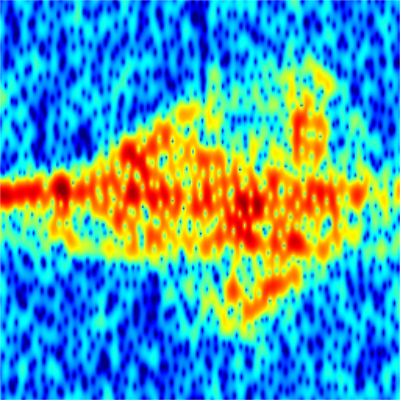}
			\end{overpic}	
		\end{minipage}%
		\begin{minipage}[t]{0.6\linewidth}
			\centering
			\begin{overpic}[width=0.298\textwidth]%
				{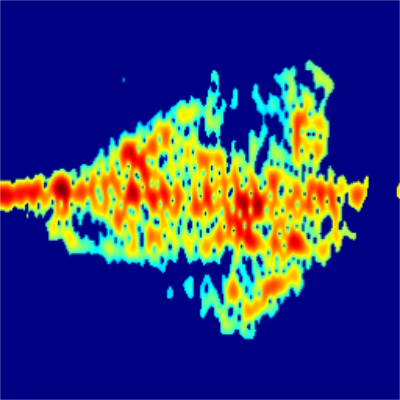}
			\end{overpic}
			\begin{overpic}[width=0.298\textwidth]%
				{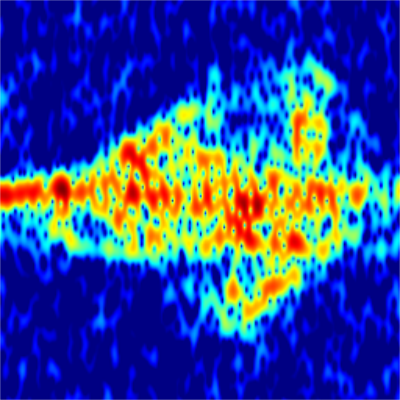}
			\end{overpic}
			\begin{overpic}[width=0.298\textwidth]%
				{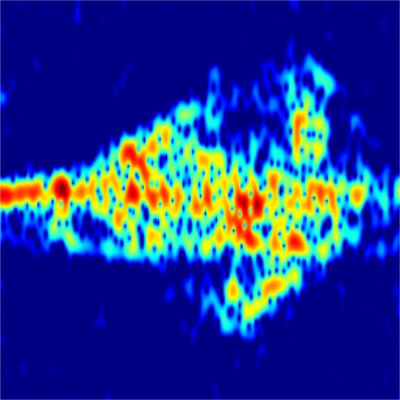}
			\end{overpic}
		\end{minipage}
		\begin{minipage}[t]{0.190\linewidth}
			\centering
			\begin{overpic}[width=0.945\textwidth]%
				{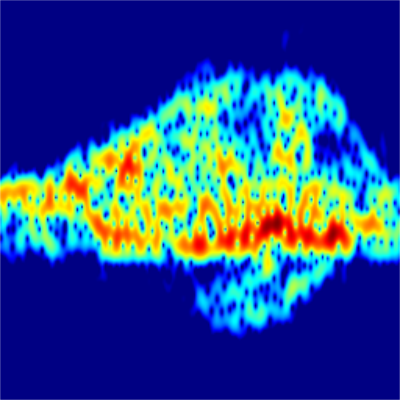}
			\end{overpic}		
		\end{minipage}\\
	\end{minipage}
	\caption{Qualitative comparison of the radar denoising results using the proposed adversarial framework in comparison to classical approaches such as CFAR and Gamma-correction. The classical approaches hyperparameters are tuned for each test subject separately. In contrast, the trained adversarial framework can generalize to all test subjects on different SNR levels.}
	\label{fig:res}
	\vspace{-5mm}
\end{figure*}
The proposed network is trained on 15 subjects and tested on the remaining 7 subjects. In training every clean spectrum is passed with 3 different SNR levels to the network, see Fig.~\ref{fig:awgn}. To have a total of 15$\times$360$\times$3$\,\approx\,$16,000 images for training. The training of the model takes 48 hours on a single NIVIDIA Titan X GPU for 200 epochs. However, the model testing inference time is only \unit[100]{ms}. To evaluate the proposed architecture performance, the real noisy $\boldsymbol{\mu}$-Ds of the remaining 7 subjects are denoised ($7\times 360 = 2520$ images). Furthermore, a comparison with other classical denoising techniques is applied to the test set. The first denoising approach used for comparison is based on a 2D CA-CFAR proposed in \cite{kronauge2013cfar2d}, where an adaptive noise threshold is calculated based on a false alarm probability and an average power of neighboring elements with an appropriate guard-band. The CFAR detector hyperparameters are tuned based on the visual aspect with suitable guard-band and probability of false alarm for each subject in the 7 subjects. The second approach is based on Gamma-correction described in \cite{poynton1998rehabilitation}, where a non-linear mapping based on a factor $\gamma$ is applied to suppress the noise. In this approach, again the factor $\gamma$ is tuned manually to an acceptable noise suppression behavior for each subject. Furthermore, a quantitative comparison is applied via different evaluation metrics such as Structural Similarity Index (SSIM) \cite{wang2004ssim}, Peak Signal to Noise Ratio (PSNR), the Mean Squared Error (MSE) and Visual Information Fidelity (VIF) \cite{sheikh2006image}. Due to the lack of perfectly paired clean spectra, the quantitative metrics of the denoised results were calculated in comparison to clean spectra from the same subjects and same gait orientation. As such, the calculated metrics illustrates the general performance differences between the approaches rather than a precise measure of the denoising accuracy.
%\vspace{-3mm}
\section{Results}
\begin{table}[!t]
	\vspace{1mm}
	\caption{Quantitative comparison of different radar denoising techniques.\label{tab:res}}
	\centering
	\setlength\arrayrulewidth{0.05pt}
	\small
	\bgroup
	\def\arraystretch{1.75}
	%	\begin{adjustbox}{width=0.48\textwidth}
	\resizebox{\columnwidth}{!}{%
		\begin{tabular}{r|cccc}
			\hline\hline
			Model & SSIM & PSNR(dB) & MSE & VIF\\
			\hline
			GC & 0.6920 & 9.55 & 7275.3 & 0.0373\\
			CFAR & 0.6884 & 9.57 & 7213.1 & 0.04367\\
			%		Style-content & 0.9046 & 24.12 & 282.8 & 0.4105 & 0.9435 & 0.2432\\
			cGAN & \textbf{0.7231} & \textbf{10.54} & \textbf{5792.2} & \textbf{0.0597}
			\\
			\hline \hline
		\end{tabular}
		%	\end{adjustbox}
	}
	\egroup
	\vspace{-5mm}
\end{table}
From a qualitative visual inspection in Fig.~\ref{fig:res}, the proposed architecture can suppress noise corrupting the background or the inside of the spectrum on different SNR levels, while keeping the vital components of the $\boldsymbol{\mu}$-D spectra. For the CFAR technique, the noise in the background is suppressed. However, the noise components in the vital $\boldsymbol{\mu}$-D part is not suppressed. This is due to the number of guard cells, which is tuned to a safe margin that can keep the threshold calculation localized to the background noise and as far as possible from the vital components in the spectrum. The Gamma-correction can remove the background noise, but with the expense of removing low-power spectrum components for low and intermediate SNR cases. This unwanted behavior can highly affect post-processing performance, such as activity recognition or human identification. 

A quantitative comparison shown in Table~\ref{tab:res} is used to give an impression of the denoising performance of the presented techniques with the clean target dataset from the same subjects. Across all four quantitative metrics, the performance of the proposed cGAN framework proved to enhance the $\boldsymbol{\mu}$-D spectra denoising in comparison to CFAR and Gamma-correction. Moreover, the proposed architecture can generalize over all unseen test subjects unlike the traditional techniques used for comparison which are tuned for each subject separately.
%\vspace{-3mm}
\section{Conclusion}
\vspace{-1mm}
In this paper, a cGAN-based radar is introduced for denoising of $\boldsymbol{\mu}$-D spectrum. The proposed framework consists of a U-Net generator trained in competition with a discriminator to translate a noisy source domain spectrum to a denoised clean spectrum. The network combines adversarial losses and non-adversarial losses to insure realistic spectrum reconstruction. The $\boldsymbol{\mu}$-D of 22 subjects with different heights, weights and genders walking on a treadmill were collected in one clean and one noisy experiment. Then 15 subjects were used for training and 7 subjects for testing. To insure alignment during training and improve the network generalization capability, an AWGN with different SNR is added to the clean target spectrum to generate aligned noisy $\boldsymbol{\mu}$-D. Finally, the network is tested on traditional real noisy $\boldsymbol{\mu}$-D and the results were compared with two techniques: 2D-CFAR and Gamma correction. The proposed architecture proved to be better than other techniques on qualitative and quantitative basis.

For future work, the modified cGAN architecture should be trained on the $\boldsymbol{\mu}$-Ds of different human activities such as falling, sitting, etc. This will enable the network to generalize on any type of activity or SNR level. Accordingly, the denoised output should be tested on different post-processing classification problems. 
\bibliographystyle{IEEEtran}

%\begin{thebibliography}{1}
%\end{thebibliography}

\end{document}

%% file: fig/cleanMD_new.tex
\begin{tikzpicture}

\begin{axis}[%
width=4.521in,
height=3.566in,
at={(0.758in,0.481in)},
scale only axis,
point meta min=-22.525091228843,
point meta max=17.474908771157,
axis on top,
xmin=16.7811864907816,
xmax=17.8371982263316,
xlabel style={font=\color{white!15!black}},
xlabel={\LARGE Time [s]},
ymin=-5.53824032,
ymax=6.00758272,
ytick style={draw=none},
xtick style={draw=none},
yticklabel style = {font=\Large},
xticklabel style = {font=\Large, yshift=-0.2cm},
ylabel style={font=\color{white!15!black}, yshift=-0.2cm},
ylabel={\LARGE Velocity [m/s]},
axis background/.style={fill=white},
legend style={legend cell align=left, align=left, draw=white!15!black}
]
\addplot [forget plot] graphics [xmin=16.7811864907816, xmax=17.8371982263316, ymin=-5.53824032, ymax=6.00758272] {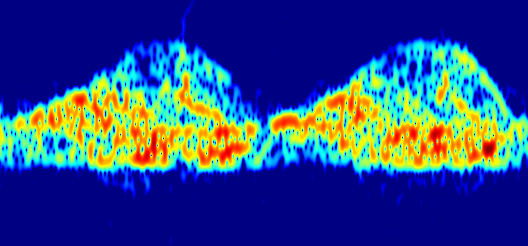};
\end{axis}

\begin{axis}[%
width=5.833in,
height=4.375in,
at={(0in,0in)},
scale only axis,
xmin=0,
xmax=1,
ymin=0,
ymax=1,
axis line style={draw=none},
ticks=none,
axis x line*=bottom,
axis y line*=left,
legend style={legend cell align=left, align=left, draw=white!15!black}
]
\end{axis}
\end{tikzpicture}%

%% file: fig/noisyMD_new.tex
\begin{tikzpicture}

\begin{axis}[%
width=4.521in,
height=3.566in,
at={(0.758in,0.481in)},
scale only axis,
point meta min=-41.2614976738841,
point meta max=3.73850232611587,
axis on top,
xmin=16.6371848904793,
xmax=17.6511961592745,
xlabel style={font=\color{white!15!black}},
xlabel={\LARGE Time [s]},
ymin=-5.53824032,
ymax=6.00758272,
ytick style={draw=none},
xtick style={draw=none},
yticklabel style = {font=\Large},
xticklabel style = {font=\Large, yshift=-0.2cm},
ylabel style={font=\color{white!15!black}, yshift=-0.2cm},
ylabel={\LARGE Velocity [m/s]},
axis background/.style={fill=white},
legend style={legend cell align=left, align=left, draw=white!15!black}
]
\addplot [forget plot] graphics [xmin=16.6371848904793, xmax=17.6511961592745, ymin=-5.53824032, ymax=6.00758272] {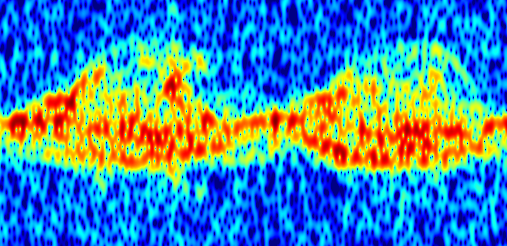};
\end{axis}

\begin{axis}[%
width=5.833in,
height=4.375in,
at={(0in,0in)},
scale only axis,
xmin=0,
xmax=1,
ymin=0,
ymax=1,
axis line style={draw=none},
ticks=none,
axis x line*=bottom,
axis y line*=left,
legend style={legend cell align=left, align=left, draw=white!15!black}
]
\end{axis}
\end{tikzpicture}%

%% file: fig/noiseAnalysis.tex
\begin{tikzpicture}

\begin{axis}[%
width=4.23in,
height=3.353in,
at={(0.71in,0.453in)},
scale only axis,
xmin=-103.6,
xmax=107.6,
xlabel style={font=\color{white!15!black}},
xlabel={\Large Power [dB]},
ymin=0,
ymax=0.045,
yticklabel style = {font=\large},
xticklabel style = {font=\large},
ylabel style={font=\color{white!15!black}},
ylabel={\Large Probability Density Function},
axis background/.style={fill=mycolor2},
legend style={legend cell align=left, align=left, draw=white!15!black, fill=mycolor2}
]

\addplot[ybar interval, fill=red, fill opacity=0.6, draw=black, area legend] table[row sep=crcr] {%
x	y\\
-94	1.05429636111065e-09\\
-92.08	0\\
-90.16	0\\
-88.24	0\\
-86.32	1.05429636111066e-09\\
-84.4	0\\
-82.48	4.2171854444426e-09\\
-80.56	2.1085927222213e-09\\
-78.64	7.38007452777455e-09\\
-76.72	3.16288908333195e-09\\
-74.8	4.2171854444426e-09\\
-72.88	1.58144454166599e-08\\
-70.96	2.21402235833235e-08\\
-69.04	2.21402235833238e-08\\
-67.12	3.69003726388727e-08\\
-65.2	6.53663743888603e-08\\
-63.28	9.59409688610691e-08\\
-61.36	1.47601490555491e-07\\
-59.44	2.31945199444344e-07\\
-57.52	3.70058022749838e-07\\
-55.6	5.33473958721989e-07\\
-53.68	8.72957386999621e-07\\
-51.76	1.28940444963832e-06\\
-49.84	1.98734864069357e-06\\
-47.92	3.15234611972084e-06\\
-46	4.97838741716449e-06\\
-44.08	7.46336394030232e-06\\
-42.16	1.19852410331059e-05\\
-40.24	1.82171868236309e-05\\
-38.32	2.86083317587376e-05\\
-36.4	4.4468111918925e-05\\
-34.48	6.8994208167442e-05\\
-32.56	0.000107826051739869\\
-30.64	0.000167595166747593\\
-28.72	0.000259059593259387\\
-26.8	0.000401080693175519\\
-24.88	0.000619062791613317\\
-22.96	0.000952616857156055\\
-21.04	0.00145936651171661\\
-19.12	0.00222968393077666\\
-17.2	0.00337550903047713\\
-15.28	0.00505812701384437\\
-13.36	0.00746387105685198\\
-11.44	0.0107772883574379\\
-9.52000000000001	0.0151265971675585\\
-7.60000000000001	0.0204450478759992\\
-5.68000000000001	0.0263498244197444\\
-3.76000000000001	0.0321304196122998\\
-1.84	0.0368689134897536\\
0.0799999999999983	0.0398487767247967\\
2	0.0408007546783579\\
3.92	0.0399560513795397\\
5.84	0.0378015219105687\\
7.75999999999999	0.0348059623113011\\
9.67999999999999	0.0312949213750105\\
11.6	0.027499524058572\\
13.52	0.0235547498165059\\
15.44	0.0196235296084582\\
17.36	0.015870338938244\\
19.28	0.0124623132993976\\
21.2	0.00952419703736527\\
23.12	0.00708598172073182\\
25.04	0.00515377067113161\\
26.96	0.00366975365619587\\
28.88	0.00256702186595943\\
30.8	0.00176511456503142\\
32.72	0.0012016279517977\\
34.64	0.000806351160090085\\
36.56	0.000537802899580713\\
38.48	0.00035463682987587\\
40.4	0.000232085420860372\\
42.32	0.000152326846845988\\
44.24	9.83985336788188e-05\\
46.16	6.39578344504169e-05\\
48.08	4.13874579517594e-05\\
50	2.65292593346275e-05\\
51.92	1.74549305545478e-05\\
53.84	1.09098587447731e-05\\
55.76	7.1038488811635e-06\\
57.68	4.66315280519244e-06\\
59.6	3.01950477822092e-06\\
61.52	1.87559322641583e-06\\
63.44	1.24512400247169e-06\\
65.36	7.75962121777432e-07\\
67.28	5.17659513305333e-07\\
69.2	2.93094388388758e-07\\
71.12	2.39325273972119e-07\\
73.04	1.24406970611058e-07\\
74.96	8.75065979721833e-08\\
76.88	4.84976326110902e-08\\
78.8	5.06062253333108e-08\\
80.72	2.53031126666558e-08\\
82.64	1.26515563333277e-08\\
84.56	1.26515563333279e-08\\
86.48	4.21718544444257e-09\\
88.4	2.10859272222131e-09\\
90.32	3.16288908333197e-09\\
92.24	0\\
94.16	3.16288908333197e-09\\
96.08	2.10859272222128e-09\\
98	2.10859272222128e-09\\
};

\addplot [color=mycolor1, line width=2.0pt]
  table[row sep=crcr]{%
-92.8	1.05429636111065e-09\\
-90.88	0\\
-88.96	0\\
-87.04	0\\
-85.12	1.05429636111066e-09\\
-83.2	0\\
-81.28	4.2171854444426e-09\\
-79.36	2.1085927222213e-09\\
-77.44	7.38007452777455e-09\\
-75.52	3.16288908333195e-09\\
-73.6	4.2171854444426e-09\\
-71.68	1.58144454166599e-08\\
-69.76	2.21402235833235e-08\\
-67.84	2.21402235833238e-08\\
-65.92	3.69003726388727e-08\\
-64	6.53663743888603e-08\\
-62.08	9.59409688610691e-08\\
-60.16	1.47601490555491e-07\\
-58.24	2.31945199444344e-07\\
-56.32	3.70058022749838e-07\\
-54.4	5.33473958721989e-07\\
-52.48	8.72957386999621e-07\\
-50.56	1.28940444963832e-06\\
-48.64	1.98734864069357e-06\\
-46.72	3.15234611972084e-06\\
-44.8	4.97838741716449e-06\\
-42.88	7.46336394030232e-06\\
-40.96	1.19852410331059e-05\\
-39.04	1.82171868236309e-05\\
-37.12	2.86083317587376e-05\\
-35.2	4.4468111918925e-05\\
-33.28	6.8994208167442e-05\\
-31.36	0.000107826051739869\\
-29.44	0.000167595166747593\\
-27.52	0.000259059593259387\\
-25.6	0.000401080693175519\\
-23.68	0.000619062791613317\\
-21.76	0.000952616857156055\\
-19.84	0.00145936651171661\\
-17.92	0.00222968393077666\\
-16	0.00337550903047713\\
-14.08	0.00505812701384437\\
-12.16	0.00746387105685198\\
-10.24	0.0107772883574379\\
-8.32000000000001	0.0151265971675585\\
-6.40000000000001	0.0204450478759992\\
-4.48000000000001	0.0263498244197444\\
-2.56	0.0321304196122998\\
-0.640000000000003	0.0368689134897536\\
1.28	0.0398487767247967\\
3.2	0.0408007546783579\\
5.12	0.0399560513795397\\
7.04	0.0378015219105687\\
8.95999999999999	0.0348059623113011\\
10.88	0.0312949213750105\\
12.8	0.027499524058572\\
14.72	0.0235547498165059\\
16.64	0.0196235296084582\\
18.56	0.015870338938244\\
20.48	0.0124623132993976\\
22.4	0.00952419703736527\\
24.32	0.00708598172073182\\
26.24	0.00515377067113161\\
28.16	0.00366975365619587\\
30.08	0.00256702186595943\\
32	0.00176511456503142\\
33.92	0.0012016279517977\\
35.84	0.000806351160090085\\
37.76	0.000537802899580713\\
39.68	0.00035463682987587\\
41.6	0.000232085420860372\\
43.52	0.000152326846845988\\
45.44	9.83985336788188e-05\\
47.36	6.39578344504169e-05\\
49.28	4.13874579517594e-05\\
51.2	2.65292593346275e-05\\
53.12	1.74549305545478e-05\\
55.04	1.09098587447731e-05\\
56.96	7.1038488811635e-06\\
58.88	4.66315280519244e-06\\
60.8	3.01950477822092e-06\\
62.72	1.87559322641583e-06\\
64.64	1.24512400247169e-06\\
66.56	7.75962121777432e-07\\
68.48	5.17659513305333e-07\\
70.4	2.93094388388758e-07\\
72.32	2.39325273972119e-07\\
74.24	1.24406970611058e-07\\
76.16	8.75065979721833e-08\\
78.08	4.84976326110902e-08\\
80	5.06062253333108e-08\\
81.92	2.53031126666558e-08\\
83.84	1.26515563333277e-08\\
85.76	1.26515563333279e-08\\
87.68	4.21718544444257e-09\\
89.6	2.10859272222131e-09\\
91.52	3.16288908333197e-09\\
93.44	0\\
95.36	3.16288908333197e-09\\
97.28	2.10859272222128e-09\\
};

\end{axis}
\end{tikzpicture}%